\documentclass[runningheads]{llncs}
\usepackage{graphicx}

\usepackage{tikz}
\usepackage{comment}
\usepackage{amsmath,amssymb} 
\usepackage{color}

\usepackage{times}
\usepackage{epsfig}
\usepackage[ruled,vlined]{algorithm2e}
\usepackage{float}
\usepackage{array}
\usepackage{makecell}
\usepackage{multirow}
\usepackage{changepage}
\usepackage{caption}
\usepackage{subcaption}
\newcommand{\ie}{\textit{i}.\textit{e}., }
\newcommand{\eg}{\textit{e}.\textit{g}. }

\SetKw{KwBy}{by}
\SetKw{KwFrom}{from}
\SetKw{KwUse}{use}

\renewcommand{\lastandname}{\unskip,}

\begin{document}
\pagestyle{headings}
\mainmatter
\def\ECCVSubNumber{5875}  

\title{DA-NAS: Data Adapted Pruning for Efficient Neural Architecture Search} 


%
\author{Xiyang Dai \and  \quad Dongdong Chen \and \quad Mengchen Liu  \and \quad Yinpeng Chen \and \quad Lu Yuan}
\authorrunning{X. Dai et al.}
%
\institute{Microsoft}
\maketitle

\begin{abstract}
Efficient search is a core issue in Neural Architecture Search (NAS). It is difficult for conventional NAS algorithms to directly search the architectures on large-scale tasks like ImageNet. In general, the cost of GPU hours for NAS grows with regard to training dataset size and candidate set size. One common way is searching on a smaller proxy dataset (\eg, CIFAR-10) and then transferring to the target task (\eg, ImageNet). These architectures optimized on proxy data are not guaranteed to be optimal on the target task. Another common way is learning with a smaller candidate set, which may require expert knowledge and indeed betrays the essence of NAS. In this paper, we present \emph{DA-NAS} that can directly search the architecture for large-scale target tasks while allowing a large candidate set in a more \emph{efficient} manner. Our method is based on an interesting observation that the learning speed for blocks in deep neural networks is related to the difficulty of recognizing distinct categories. We carefully design a progressive data adapted pruning strategy for efficient architecture search. It will quickly trim low performed blocks on a subset of target dataset (\eg, easy classes), and then gradually find the best blocks on the whole target dataset. At this time, the original candidate set becomes as compact as possible, providing a faster search in the target task. Experiments on ImageNet verify the effectiveness of our approach. It is $\textbf{2}\boldsymbol{\times}$ faster than previous methods while the accuracy is currently state-of-the-art, at \textbf{76.2\%} under small FLOPs constraint. It supports an argument search space (\ie, more candidate blocks) to efficiently search the best-performing architecture.

\end{abstract}

\section{Introduction}

Neural Architecture Search (NAS) has a great impact by automating neural network architecture design. The architecture is optimized for accuracy and efficiency (especially latency) under the constraints (\eg, FLOPs, latency, memory). Recently, NAS has demonstrated the success in various deep learning tasks, such as image classification \cite{guo2019single,liu2018darts,wu2019fbnet}, detection \cite{chen2019detnas} and segmentation \cite{liu2019auto,nekrasov2019fast}.

Despite the remarkable results, conventional NAS algorithms~\cite{Tan_2019_CVPR,liu2018progressive,nasnet_2018_CVPR} is prohibitively computation-intensive, especially directly on a large-scale task (\eg, ImageNet~\cite{deng2009imagenet}), which makes it difficult for making paretical industry impact. As a result, one common way is to utilize a smaller proxy data (\eg, CIFAR-10) for searching, and then transfer to the large-scale target task (\eg, ImageNet) \cite{liu2018darts,liang2019darts+,liu2018progressive,real2019regularized}. Due to the domain gap (\eg, resolution, class number) between proxy data and target task, these blocks optimized on proxy data are not guaranteed to be optimal on the target task, especially when taking accuracy and resource constraint into consideration. Thus, \emph{directly} searching on the target dataset is essential to NAS.

Another common way is searching with a smaller candidate set \cite{guo2019single,cai2018proxylessnas,wu2019fbnet}, which highly relies on the expert knowledge and indeed betrays the essence of NAS. In addition, too few candidate blocks are not beneficial to find a best-performing architecture under search constraints (\eg, FLOPs, latency). Thus, an argument search space with more candidate blocks is always encouraged to boost the performance of NAS.

In this paper, we propose a simple and effective solution to the aforementioned limitations, called \emph{DA-NAS}, which can directly search the architecture for large-scale target tasks in a more efficient manner, while allowing a large candidate set. The solution is based on our observation that the learning speed of blocks in deep neural networks is varied in different classes (for classification task). The blocks are learnt much faster in easy classes than in difficult classes. Besides, our study indicates that the performance of blocks in easy classes converges very quickly at the early training stage but needs more time in difficult ones. The discovery motivates a new data adapted pruning for NAS, which starts the search on a subset of target task (\eg, easy classes), and gradually trims low performed blocks as the size of subset increases until we find the best blocks on the whole target dataset. To build the strategy, we may be able to group classes based on the easiness, and feed them progressively to reduce the computation cost.

We formulate NAS as a block-level pruning process, which is different from recent ProxylessNAS~\cite{cai2018proxylessnas} that adopts a path-level pruning. Specifically, we directly train a supernet~\cite{xie2018snas}, an over-parameterized network that contains all candidate paths. In the beginning of training, we train it on a subset of target task (only containing easy classes). During training, we progressively prune low performed blocks from our candidate set until we get a compact candidate set for searching on the whole dataset of target task. We consider a loss function with cost constraint which helps find an optimal architecture under search constraint (\eg, FLOPs).

Comprehensive experiments and comparisons to existing methods demonstrate that DA-NAS can find an optimal architecture $2\times$ faster and is also capable of finding a current best small FLOPs architecture at 76.2\% on ImageNet within a highly complex search space (involving inverted residual block, shuffle block, squeeze-and-excitation block and more).

Our contributions can be summarized as follows:
\begin{itemize}
  \item DA-NAS is the first NAS algorithm that shows a close connection between block pruning and dataset scheduling. To our best knowledge, it is the first work to study the relationship between network learning and training data for NAS.  
  \item We propose a progressive block-level pruning perspective for NAS, according to data adaption. It can search architecture on the large-scale target task much faster, and effectively enlarge the search space to achieve state-of-the-art performance.
  \item The DA-NAS is convenient to use for various needs. It enables cost constraint in search, which is beneficial to practical industry impact. The inherent idea is also generalized to other tasks, like key-point localization.
\end{itemize}

\section{Related Work}
\paragraph{Efficient Network.} Since the need of delopying deep neural networks into real application systems is increasing, efficient network has drawn a lot of attention from both academia and industry. Existing research about efficient network is often done from two broad aspects:  efficient network structure design \cite{howard2017mobilenets,sandler2018mobilenetv2,ma2018shufflenet,zhang2018shufflenet}, or pruning/quantizing one given network structure \cite{he2017channel,zhou2017incremental,han2015deep}. In this paper, we focus on the former problem. For efficient network architecture design, many interesting approaches have been proposed. For example, Xception \cite{chollet2017xception} proposes to decompose one normal convolution layer into one depth-wise and one point-wise convolutional layer, which is able to significantly reduce the computation FLOPs. Based on this design scheme, a lot of efficient networks have been further designed, such as MobileNet \cite{howard2017mobilenets,sandler2018mobilenetv2,howard2019mbnetv3}, ShuffleNet~\cite{zhang2018shufflenet,ma2018shufflenet}. Despite their success, designing such an efficient network is not that easy and can only be done by experts.    

\paragraph{Neural Architecture Search.} Recently, NAS has drawn surging interests that study how to automatically design a better and efficient network structure with machine learning algorithms. Based on the searching strategy, existing NAS methods can also be roughly divided into two categories, \ie, searching an efficient operator block~\cite{liu2018darts,liang2019darts+,pham2018efficient,yan2019hm,Tan_2019_CVPR,real2019regularized} from scratch, or finding an optimal operator combination from a pre-defined efficient operator search space~\cite{guo2019single,wu2019fbnet,cai2018proxylessnas,cai2019once}. Compared to the former category, the latter category of approaches leverage a lot of design priors from human experts, so it is relatively easier to find an optimal network architecture. Our method belongs to the latter category. By contrast to existing methods whicj often regard data scheduling and architecture search as two independent parts, our method is the first that shows a close connection between both. By leveraging a new and efficient data scheduling mechanism, a progressive block-level search space pruning algorithm is further proposed. Our method is demonstrated to be more efficient and can search a better architecture given the same searching time.

\section{Understanding Network Training Process}
\label{sec:analysis}

In this section, we analyze the relationship between the performance of deep neural networks and the training data. These interesting observations will inspire our data adapted pruning for efficient neural architecture search.
\newline



\noindent \textbf{Observation 1.} \textit{There exists some classes that are easy to learn (easy classes) while some classes are harder to learn (hard classes).}
\newline

\begin{figure}
\begin{center}
\includegraphics[width=0.6\linewidth]{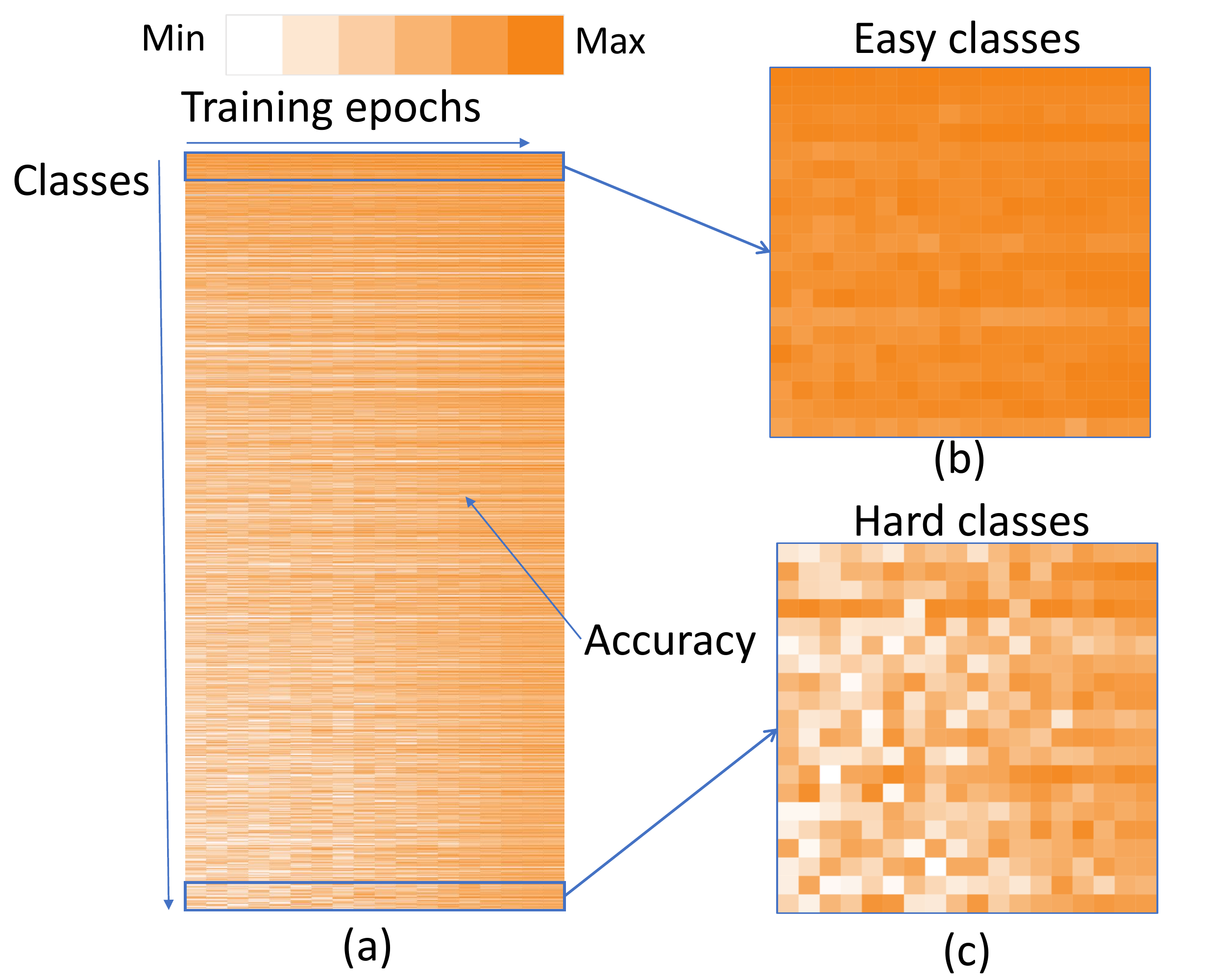}
\caption{Comparing between learning curves of different classes.}
\label{fig:data2}
\end{center}
\vspace{-0.3cm}
\end{figure}

We start our analysis from exploring a typical network training process, \ie, ResNet-34 trained on ImageNet. A matrix shown in Figure~\ref{fig:data2} (a) visualizes the accuracy of distinct classes varies with more training iterations. Each row is certain a class and each column is training time (epoch $=10,20,...,180$). The value at each grid denotes the accuracy for each class. For a better visualization, we sort the matrix rows by row-wise variance of a matrix, namely, the variance of the accuracies of recognizing every class from each training epoch.

We find that the accuracy of some classes quickly converges and achieves its maximum (Figure.~\ref{fig:data2} (b)); while some other classes gradually increase / fluctuate in the training (Figure.~\ref{fig:data2} (c)). It indicates the learning speed for every class is different. Thus, we can group classes based on their easiness and feed them heuristically into training, following small-to-large data scheduling. Meanwhile, search using fewer categories at the beginning is a considerably easier task than the search using all categories in the end. This helps us progressively trim low performed blocks in the search space to reduce the search cost.
\newline


\noindent \textbf{Observation 2.} \textit{Neurons to recognize easier classes converge more quickly at the early training stage and their performance remains stable in the remaining training process. Neurons for hard classes need more time to be fully trained.}
\newline

\begin{figure}
\begin{center}
\includegraphics[width=1\linewidth]{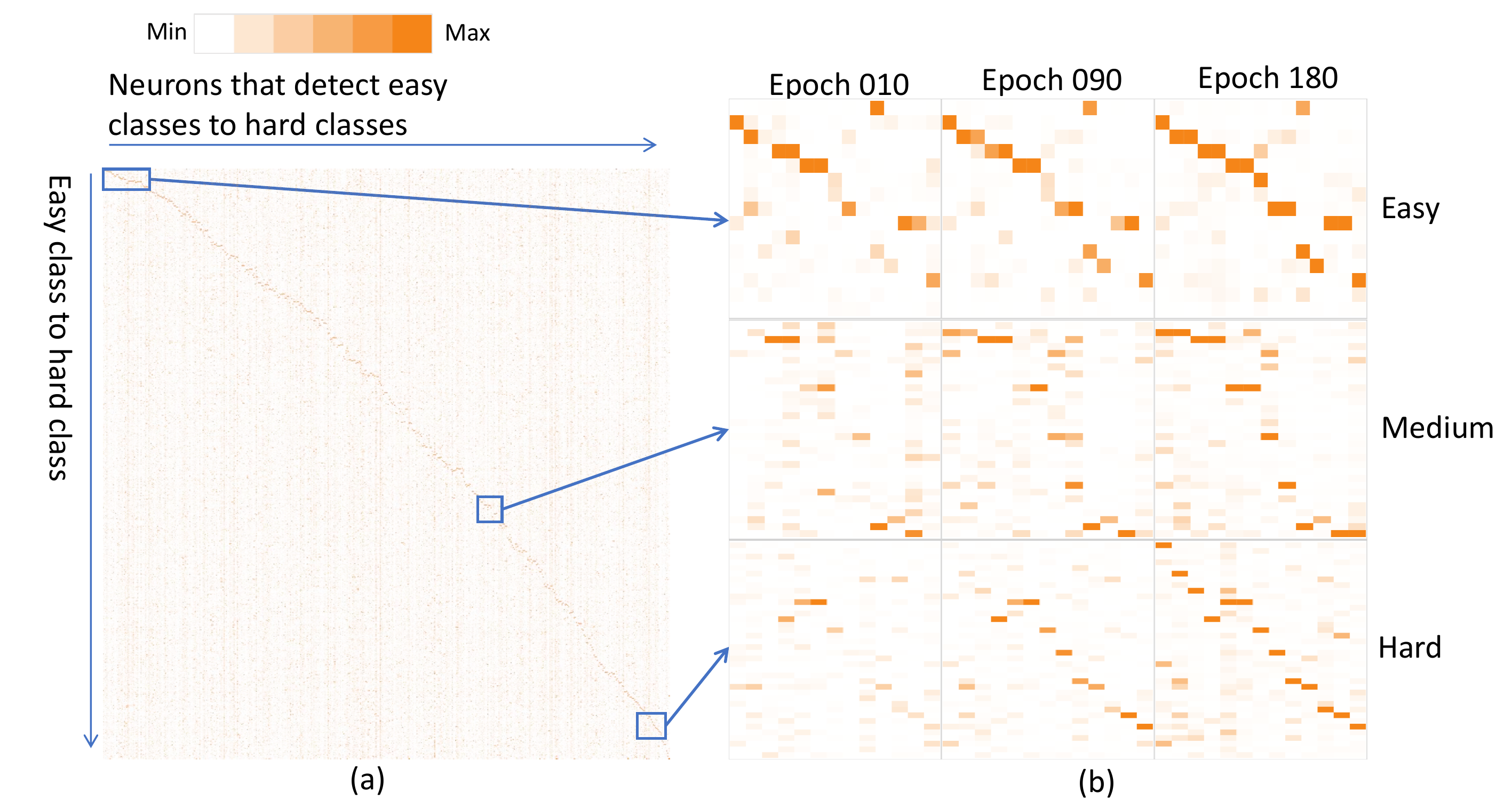}
\caption{Comparing neuron responses for easy and hard classes at different timestamps in the training.}
\label{fig:visualization_neuron}
\end{center}
\vspace{-0.3cm}
\end{figure}

We study why the learning speed of different classes are not equal (in Observation 1). One possible reason is that various learning speed of neurons (\ie, convolution kernels) makes learning speed of distinct classes different, since the final prediction of network is determined by a combination of neuron responses. To verify this reason, we visualize the relations between neurons and classes in increasing training epochs on Figure~\ref{fig:visualization_neuron} (b). The relation between a neuron and a class, used to distinguish the class from other classes, is computed by the neuron relevance measure proposed in~\cite{bilal2017convolutional}. Specifically, we only consider the neurons in the final convolutional layer (before FC-layer) since it represents the maximum semantics in neurons. We collect all the calculated relations into a matrix, where each row is a class and each column is a neuron. To be consistent with Figure~\ref{fig:data2}, we use the same order of classes in rows, and further sort the matrix columns by column-wise maximum value (corresponding to the class most likely to be recognized by the neuron) of a matrix. Figure~\ref{fig:visualization_neuron} (a) show the sorting result of the matrix, where the top-left block corresponds to the easiest classes and the neurons that are most likely used to recognize these classes, and the right-bottom block corresponds to the hardest classes and corresponding neurons.

Figure~\ref{fig:visualization_neuron} (a) shows such a visualization corresponding to the last epoch, where classes of different easiness levels are learned by different neurons.
More interestingly, there is a ``diagonal pattern'' that neurons are distributed in balance across all the classes, summarized in Observation 3. It indicates that \textbf{all} the classes should be involved in the training for the best performance. In other words, directly learning on a large-scale dataset should yield better accuracy than on a smaller dataset. 

To further investigate the evolutionary pattern in the neuron-class relations, we visualize the matrix at three different training epochs (epoch=$10,90,180$) and select three representative blocks (corresponding to easy, medium, and hard classes) at each epoch, shown in Figure~\ref{fig:visualization_neuron} (b). In the first row, we observe that the neurons to recognize easy classes quickly converge and their performance keep stable in subsequent training. Compared with other two rows, the neurons learn faster on easy classes (stable on epoch=10) than that on medium classes (almost stable on epoch=90), and even much faster than hard classes (almost on epoch=180). The phenomena is summarized in Observation 2.
\newline



\noindent \textbf{Observation 3.} \textit{Different architecture of networks agree on similar easy/hard classes distribution.} 
\newline

\begin{figure}
\begin{center}
\includegraphics[width=1\linewidth]{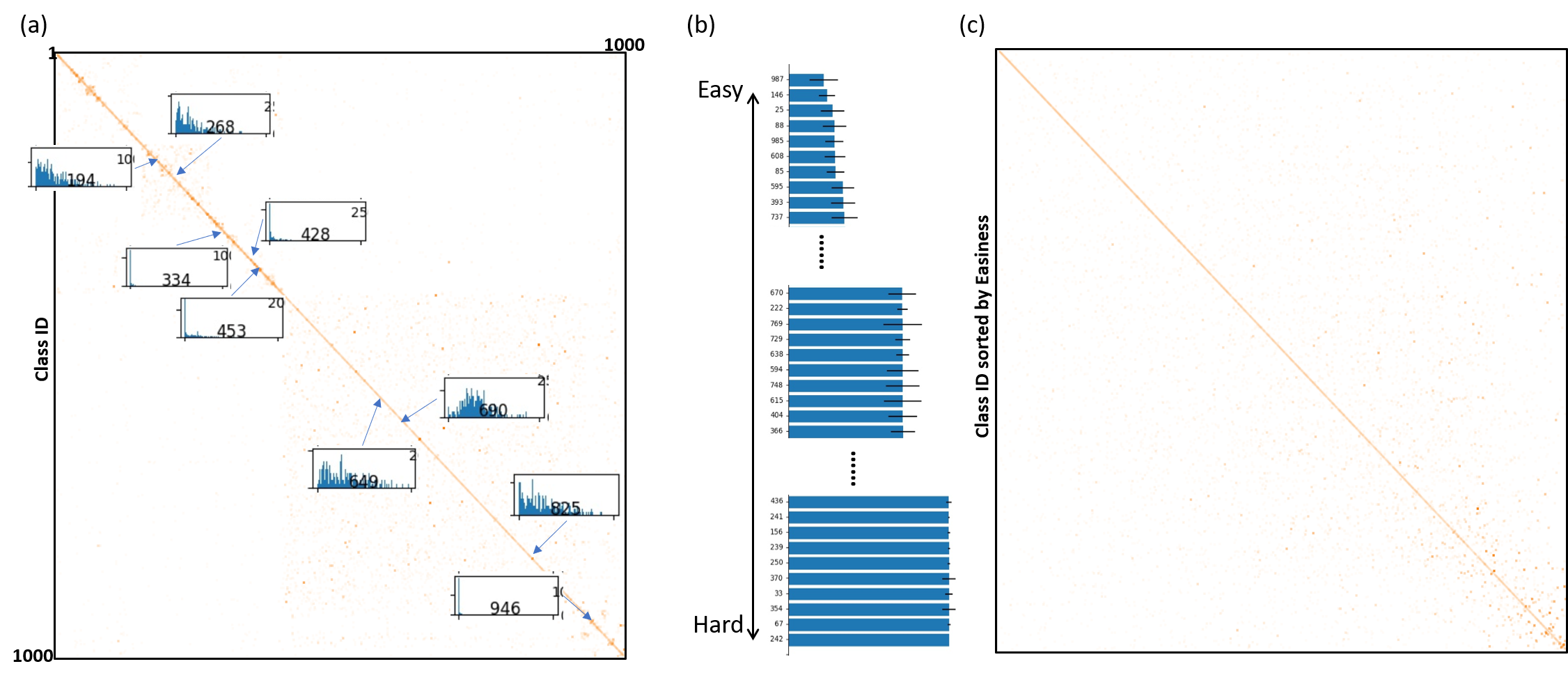}
\caption{An illustration that our defined class easiness measurement is agreed and persisted among multiple deep networks}
\label{fig:data}
\end{center}
\vspace{-1cm}
\end{figure}

To further investigate whether above observation 1 and 2 are shared among other networks or unique for ResNet, we train and evaluate a bunch of state-of-the-art manually designed networks, including VGG~\cite{simonyan2014very}, ResNet~\cite{he2016deep}, MobileNet~\cite{howard2017mobilenets,sandler2018mobilenetv2}, ShuffleNet~\cite{zhang2018shufflenet,zhang2018shufflenet} and more. For each class, we calculate the mean and variance of the running average on \textit{easiness} histograms among multiple networks. Shown in Figure \ref{fig:data} (a), the \textit{easiness} histograms are well aligned with confusion matrix and can be used as an indicator for measuring how easy a network can distinguish a class from other classes. Shown in Figure \ref{fig:data} (b), it is true that different networks agree on the similar distribution of easiness on classes. This phenomena enables us to effectively use data while increasing searching space. Shown in Figure \ref{fig:data} (c), after sorted by easiness rank, the confusion matrix disentangles the hierarchy within classes. The mathematics definition for \textit{easiness} is introduced in the following section.

These observations inspire us to design a search space pruning strategy based on the easiness of classes and the learning status of neurons. At the beginning, we only feed in a small subset with easy classes to train all candidate blocks. Then, we can explicitly exclude unfavorable blocks that still struggle to learn or perform worse than others in easy classes. By sequentially feeding more hard classes until all the classes finally (the whole dataset), we can progressively trim low performed blocks step by step, as shown in Algorithm~\ref{alg}.

\section{Data Adapted NAS}

Based on the analysis in Section.~\ref{sec:analysis}, the class easiness $E_c$ for a class $c$ is defined as: 
\begin{align}
\label{eq:easy}
    &P_i = \hat{N}(d_i) \quad for \quad d_i \in \mathcal{D} \\
    &E_{c,i} = -\sum P_i \log(P_i) \quad if \quad d_i \in c \\
    &E_c = \underset{i}{\mathcal{H}}(E_{c,i})
\end{align}
where, $\mathcal{D}$ is the training dataset (\eg, ImageNet), $\hat{N}$ is a network (\eg, ResNet-34) fine-tuned on the dataset, and $E_c$ is represented as a histogram $\mathcal{H}(\cdot)$ of entropy values of all the samples belonging to class $c$. For each training sample, since it is optimized towards a one-hot vector of the ground truth, the entropy of the network's output represents the effort of a network taking to distinguish the ground truth class from others. Then the histogram of these entropy values of a class represents the trend of easiness of all samples that belong to this class. As shown in Figure \ref{fig:data}(a), the proposed measurement is well aligned with the confusion matrix and can be used to explore the trend of easiness on all classes and all training samples. Finally, for each class, we calculate the mean of on easiness histograms among multiple networks as our final data adaptation strategy.

\subsection{Expanding Search Space}
\begin{figure} [t]
\begin{center}
\includegraphics[width=1\linewidth]{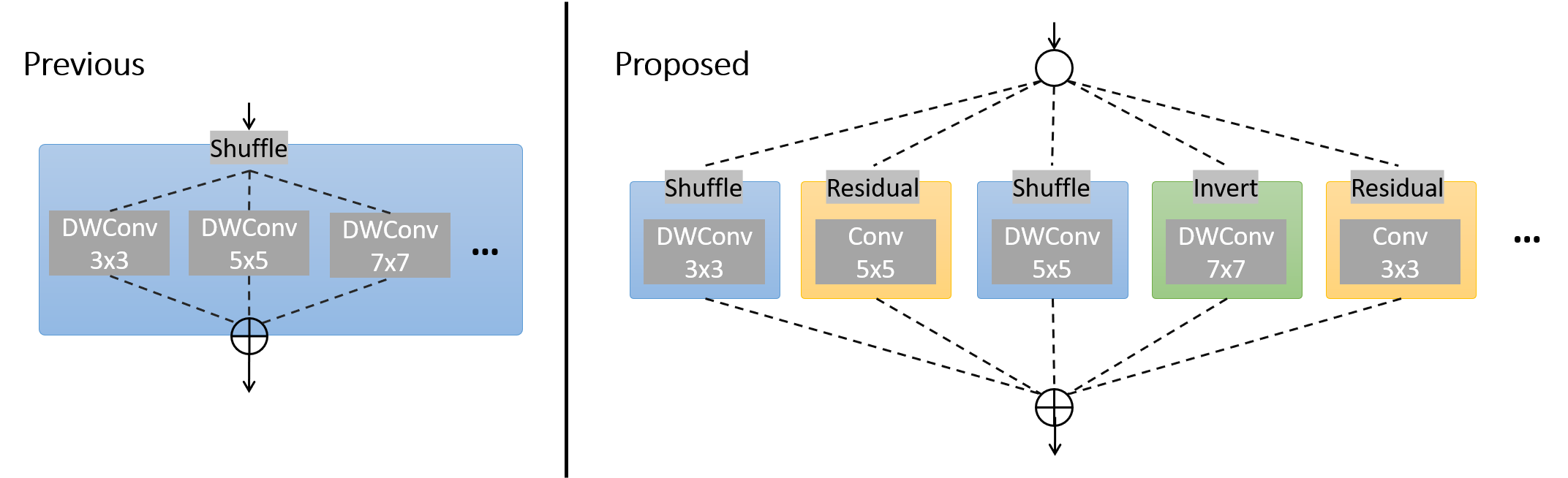}
\caption{Our proposed search space contains combinations of blocks and operators.}
\label{fig:arch}
\end{center}
\end{figure}

In order to search any possible architecture in a search space, an over-parametered super network needs to be built first. Previous methods first define a set of candidate operators (\eg, $3\times 3$ or $5\times 5$ depth-wise convolutions) $\mathcal{O} = \{o_1,\dots, o_k\}$ and build the supernet layer by layer. Such approach largely limits all possible combinations of network architectures, and makes the micro-architecture design critical to the final search result.
Compared with previous method, we allow search in a large and diverse set of candidate blocks (\eg, residual block, inverted residual block, shuffle block) $\mathcal{B} = {b_i, \dots, b_m}$ applied over candidate operators (\eg, depth-wise or normal $3\times 3$, $5\times 5$ convolutions) and effectively changed the super network to:
\begin{align}
    & N_l = \sum_j^m \sum_i^k b_j(o_i(x)) \\
    & N_{l+1} = \sum_j^m \sum_i^k b_j(o_i(N_l))
\end{align}
Figure \ref{fig:arch} shows that our design can effectively combine multiple micro-architecture build blocks and simulate popular networks.

\subsection{Searching with Constrains}
We followed the idea introduced by \cite{xie2018snas} to use Gumbel-Softmax to assist learning of the architecture:
\begin{align}
    & p \sim GumbelSoftmax(a, \tau) \\
    & N_l = \sum_j^m \sum_i^k p_{i,j}b_j(o_i(x)) 
    \label{eq:arch}
\end{align}
where $a$ is the architecture weight that we want to learn and $\tau$ is a pre-defined hyper-parameter to control the sharpness of Gumbel distribution.
To incorporate with search constrains (\eg, FLOPS or hardware latency), we also compute the expected cost of a super network by:
\begin{align}
    & C_{l} = \sum_j^m \sum_i^k p_{i,j}\mathcal{C}(b_j(o_i(x)))
    \label{eq:cost}
\end{align}
where $\mathcal{C}(\cdot)$ is a function measuring the cost of a block.

The super network is first optimized towards classification loss to find the optimal weight:
\begin{align} 
    &w = \arg\min \mathcal{L}_{cls} (N(\mathbb{D}_{train},a)).
    \label{eq:t1}
\end{align}
Then it is optimized for architecture using the modified loss function with the cost constraint:
\begin{align}
    &a = \arg\min \mathcal{L}_{cls} (N(\mathbb{D}_{val},w)) \cdot \log(\frac{C}{\beta})^\gamma
    \label{eq:t2}
\end{align}
where $\beta$ is a scaling factor designed as target cost and $\gamma$ is a factor to control the strength of incorporating cost constraints. 

\subsection{Training Strategy}
\label{sec:training}

The training of a NAS algorithm can be tedious and non-trivial. In our training strategy, we intelligently couple data scheduling and search space pruning together, as stated in previous sections. At beginning, we warm up the super network by training on only "easy" classes in order to convert each blocks into a well-trained state as fast as possible. Then we progressively add more classes from target dataset and, at same time, shrink the search space to speedup NAS to avoid falling in local minimal. The final optimal architecture is then automatically selected at the end of pruning. This strategy is further elaborated in Algorithm \ref{alg}.

We intelligently combine data adaptation with network pruning in two ways:
\begin{itemize}
    \item We progressively introduce more categories in the architecture search. The search using fewer (\eg 100) categories at the beginning is a considerably easier task than the search using all (\eg 1000) categories in the end. This helps us progressively trim low performed blocks in the search space to reduce the search cost.
    \item We also identify the easiness of the categories by voting from popular manually designed networks. We observe different network tends to learn similar "easy" categories quickly at the beginning. We start with a few "easy" categories at the beginning to speed up the convergence of supernet and hence further reduce the training cost.  
\end{itemize}

The benefit with our training strategy is three-fold:
\begin{itemize}
    \item It largely reduces the number of epochs and GPU hours required in training.
    \item It eases the difficulty of tuning hyper-parameter, especially $\tau$ in Gumbel Softmax (Equation \eqref{eq:arch}). $\tau$ controls the sample distribution and is usually critical to the final search result. Large $\tau$ turns to add randomness of sampling to help super network explore more variant of branch combinations; while lower $\tau$ trends to be more deterministic to branch selection and help super network select architecture quicker. It is necessary to balance these two factors to find optimal architecture after exploring large variants of combinations. Previous method coupled $\tau$ with learning rate and used an exponentially decaying schedule \cite{wu2019fbnet,cai2018proxylessnas}. Unlike previous methods, $\tau$ in our training strategy is set to a fixed number, as the progressively pruning of search space works similar to reducing $\tau$.
    \item It picks the final architecture more confidently. Unlike previous methods that need sampling from the final architecture distribution to find the best candidate, our final optimal architecture can be directly picked by maximizing the architecture distribution, which further saves the computational cost. 
\end{itemize}

\begin{algorithm}
\SetAlgoLined
\textbf{Input: } Training Data $D_{Train}$, Validating Data $D_{val}$, Search Space $S$, Search space pruning ratio $\pi$, Classes used per step $\delta$, Easiness of classes $E$\;
\text{$R \gets$ Sort $E$} \\
\For{$epoch \gets 1$ \KwTo $\#Epoch_{warmup}$}{
    \text{$train \gets D_{Train}$ \KwFrom $R[\delta[0]]$;} \\
    \text{Optimize Equation \eqref{eq:t1} \KwUse $train$;} \\
 }
\For{$s \gets 1$ \KwTo $\#Step$}{
    \text{$train \gets D_{Train}$ \KwFrom $R[\delta[s]]$;} \\
    \text{$val \gets D_{Val}$ \KwFrom $R[\delta[s]]$;} \\
    \For{$epoch \gets 1$ \KwTo $\#Epoch_{s}$}{
        \text{Optimize Equation \eqref{eq:t1} \KwUse $train$;} \\  
        \text{$cost \gets$ Equation \eqref{eq:cost};} \\ 
        \text{Optimize Equation \eqref{eq:t2} \KwUse $val$, $cost$;} \\    
     }
     \text{$S \gets$ Reduce $S$ \KwBy $\delta$}
 }
\Return{$S$}
\caption{NAS with Data Adapted Pruning}
\label{alg}
\end{algorithm}
\vspace{-0.5cm}

\section{Experiment}
\subsection{Setup}
\paragraph{Datasets.} 
We directly search architecture on the target dataset. For image classification task, we use the full ImageNet~\cite{deng2009imagenet} dataset. We randomly select 50 images per class from the original training set to formulate a validation set. We then use the original validation set as a test set to report final experiment results. Besides, we also want to test our DA-NAS algorithm for key-point localization task. We use full COCO~\cite{coco} key-point 2017 dataset. The original training set is divided into "trainminusminival" and "minival" for training and validation. Then the original validation set is used to report results. 

\paragraph{Search Spaces.} 
We investigate two popular search spaces widely used in previous work~\cite{howard2019mbnetv3,wu2019fbnet,cai2018proxylessnas,guo2019single} and their augments:
\begin{itemize}
    \item ``Mobile'': The search space is based on MobileNet~\cite{howard2017mobilenets,sandler2018mobilenetv2} micro-architectures. In our implementation, it contains three inverted residual blocks with expanded factor of $1,3,6$ and two depth-wise convolution operators with $3\times 3$ and $5\times 5$ kernels. 
    \item ``Shuffle'': The search space is based on ShuffleNet~\cite{ma2018shufflenet,zhang2018shufflenet} micro-architectures. In our implementation, it contains two shuffle blocks with different number of convolution layers and three depth-wise convolution operators with $3\times 3$, $5\times 5$ and $7\times 7$ kernels.
    \item ``Mobile+'': We expand "Mobile" by adding normal convolutions with $3\times 3$ and $5\times 5$ kernels into operators to increase the complexity and flops variation. 
    \item ``Shuffle+'': We expand "Shuffle" by adding normal convolutions with $3\times 3$ and $5\times 5$ kernels into operators to increase the complexity and flops variation. 
    \item ``Shuffle+Mobile'': We combine Shuffle spaces and Mobile spaces together. It is the major search space we use to find state-of-the-art architectures.
\end{itemize} 

\paragraph{Implementation Details.}
We implement our approach using Pytorch and run all experiments on a compute node with $4$ V100 GPUs. For training, the super network is first warmed up with $10$ epochs and followed by $3$ steps of search with $20$ epochs each. We use a search space pruning ratio $0.4$ and $100, 300, 600, 1000$ classes for each step respectively. For fine-tuning the searched architecture, we follow the training setup introduced by \cite{cai2018proxylessnas}, but pump up the initial learning rate to 0.5.

\begin{table}[]
\centering
\setlength{\tabcolsep}{0.8mm}{
\begin{tabular}{|l|c|c|c|l|}
\hline
\thead{Method} & \thead{Search \\ Space} & \thead{FLOPs}  & \thead{Accuracy} & \thead{Search\\Cost}\\
\hline\hline
SinglePath \cite{guo2019single} & Shuffle & 319 M  & 74.3 & \\
SinglePath (impl.) & Shuffle & 336 M  & 74.4 &  142\\
Ours & Shuffle & 325 M & 74.4 &  87$\downarrow_{39\%}$  \\
\hline
Proxyless-G \cite{cai2018proxylessnas} & Mobile & --  & 74.2 & \\
Proxyless-G (impl.) & Mobile & 420 M  & 74.6 & 399 \\
Ours & Mobile & 389 M & 74.6  &  138$\downarrow_{65\%}$  \\ 
\hline
\end{tabular}
}
\caption{Direct comparison to two popular methods with fast search speed}
\vspace{-1cm}
\label{tb:cmp1}
\end{table}

\begin{table}
\begin{center}
\begin{tabular}{|l|c|rr|}
\hline
\multicolumn{1}{|c}{\multirow{2}{*}{\thead{Method}}} & \multicolumn{1}{|c|}{\multirow{2}{*}{\thead{Search \\ Space}}} & \multicolumn{2}{c|}{\thead{~ Accuracy @ Time ~}} \\
\multicolumn{1}{|c}{} & \multicolumn{1}{|c|}{} & \multicolumn{1}{r}{~ 0.5x ~} & \multicolumn{1}{r|}{~ 1x ~} \\
\hline\hline
SinglePath (impl.) & Shuffle+ & 72.4 & 73.2 \\
Ours & Shuffle+ & 73.3 & 73.3 \\
\hline
Proxyless-G (impl.) & Mobile+ & 71.8 & 73.3 \\
Ours & Mobile+ & 73.2 & 73.3 \\
\hline
\end{tabular}
\end{center}
\caption{Restrict comparison to two popular methods on performance of searched architecture (same fine-tune setup) under constrained search time.}
\vspace{-0.5cm}
\label{tb:cmp2}
\end{table}

\subsection{Compared with State-of-the-art Methods}
We first compared with two state-of-the-art methods, ProxylessNAS~\cite{cai2018proxylessnas} and SinglePath~\cite{guo2019single}, which are claiming as fastest search methods on ImageNet. Since they report the performance and the search cost based on different criteria and hardware, in order to compare fairly, we re-implement these two methods based on public available code released by authors. As shown in Table \ref{tb:cmp1}, we are able to reproduce the reported performances. Then we run our methods with the exact same search space. It is obvious to see that our method is capable of finding competitive quality networks with much lower search cost. It significantly reduces the search cost by $39\%$ and $65\%$ respectively.

Next, to further investigate the lower bond of time cost needed for searching a proper architecture, we conduct an experiment with constrained search time and a enlarged search space. We double the operators in shuffle and mobile search spaces by introducing $3 \times 3$ convolution and $5 \times 5$ convolution. As shown in Table \ref{tb:cmp2}, our method is able to find the architecture with comparable accuracy to state-of-the-art methods by only half of the time needed in these methods.

\begin{table}
\centering
\setlength{\tabcolsep}{0.7mm}{
\begin{tabular}{|l|c|c|c|c|}
\hline
\thead{Method} & \thead{Seach \\ Dataset} & \thead{FLOPs} & \thead{Accuracy} & \thead{Seach\\Cost} \\
\hline\hline

\hline
DARTS\cite{liu2018darts} & CIFAR & 595 M & 73.1 & 96\\
SNAS\cite{xie2018snas} & CIFAR & 522 M & 72.7 & 24 \\
PNAS\cite{liu2018progressive} & CIFAR & 588 M & 74.2 & 3600 \\
NASNET-A\cite{nasnet_2018_CVPR} & CIFAR & 564 M & 74.0 & 10,000+\\
\hline
MnasNet\cite{Tan_2019_CVPR} & Imagenet & 317 M & 74.0 & 10,000+\\
FBNet\cite{wu2019fbnet} & Imagenet & 375 M & 74.9 & 216\\
Proxyless-G LL\cite{cai2018proxylessnas} & Imagenet & -- & 74.2 & 200\\
SinglePath\cite{guo2019single} & Imagenet & 319 M & 74.3 & 312\\
\hline
Ours-A & Imagenet & 323 M  & 74.3 & 138\\
Ours-B & Imagenet & 372 M  & 74.8 & 138\\
Ours-C & Imagenet & 467 M  & 76.2 & 138\\
\hline
\end{tabular}
}
\caption{Comparison to the state-of-the-art searched results on ImageNet validation set.}
\vspace{-0.5cm}
\label{tb:sota}
\end{table}

Finally, we combine ``Shuffle" and ``Mobile" search spaces together to find the state-of-the-art architecture. Table \ref{tb:sota} shows the comparison between our method with existing popular NAS approaches \cite{liu2018darts,xie2018snas,liu2018progressive,nasnet_2018_CVPR,wu2019fbnet,cai2018proxylessnas,guo2019single}. We report their search costs directly from their public papers (although some of numbers we cannot reproduce locally). Compared with methods \cite{liu2018darts,xie2018snas,liu2018progressive,nasnet_2018_CVPR} only searched on a proxy dataset (\ie, CIFAR), our method leads to a significant performance gain. It is worth noticing that the networks searched on smaller datasets suffered from sub-optimal performance when transferred to a large scale dataset. They also have difficulties in reducing FLOPs due to the fact that searching conducted on different resolutions of datasets causes different designs of architecture (such as pooling scales and number of layers). Compared with methods \cite{wu2019fbnet,cai2018proxylessnas,guo2019single,Tan_2019_CVPR} that search directly on ImageNet, our method requires the least search cost  (138 GPU hours) to find best-performing architecture with the state-of-the-art accuracy (76.2\%).

\begin{table}
\begin{center}
\begin{tabular}{|l|c|c|c|}
\hline
\thead{Method} & \thead{FLOPs} & \thead{Accuracy} & \thead{Search \\ Cost} \\
\hline\hline
Small & 319 M & 73.2 & 33\\
Small (Easy) $\rightarrow$ All & 325 M & 74.4 & 87 \\
Small (Hard) $\rightarrow$ All & 316 M & 74.1 & 87 \\
All & 327 M & 73.8 & 307\\
\hline
\end{tabular}
\end{center}
\caption{Ablation study on the effect of different data scheduling strategy.}
\vspace{-1cm}
\label{tb:ab1}
\end{table}

\subsection{Ablation Study}
We first demonstrate that our data adapted pruning is efficient. We evaluate the effects of different data scheduling on pruning: from easy classes to all classes, from hard classes to all classes, use a small subset of classes solely and use all classes directly. As shown in Table \ref{tb:ab1}, it is obvious that our "small (easy) $\rightarrow$ all" data adapted pruning is the most effective method, which is able to find the best network architecture with only $28\%$ of the time compared to searching directly on all classes. This demonstrates that "from small to all" is very important and it yields 0.9 better at top-1 accuracy because of the difference of data amount. Then, starting from easy is better than hard, it yields another 30\% compared to the improvement of using 10x more data, which is non-trivial.

\begin{table}
\begin{center}
\begin{tabular}{|l|c|c|c|}
\hline
\thead{Search Space} & \thead{FLOPs} & \thead{Accuracy} & \thead{Search \\ Cost} \\
\hline\hline
Shuffle  & 325 M & 74.4 & 87 \\
Shuffle+ & 353 M & 74.3 & 194 \\
Shuffle+Mobile & 323 M & 74.3 & 138\\
\hline
\end{tabular}
\end{center}
\caption{Ablation study on the influence of different search space on the searched architecture.}
\vspace{-0.5cm}
\label{tb:ab2}
\end{table}

Then, we show that our search space pruning is robust. We conduct experiments on three different search spaces: "Shuffle", "Shuffle+", "Shuffle+Mobile", which contain varieties of blocks and operators. As shown in Table \ref{tb:ab2}, our method is robust enough to find the optimal architectures with nearly consistent accuracy and FLOPs using different search spaces. Besides, our search cost will accordingly increase as the search space is enlarged.

\begin{table}
\begin{center}
\begin{tabular}{|cc|c|c|c|}
\hline
\thead{Scheduling} & \thead{Pruning} & \thead{FLOPs} & \thead{Accuracy} & \thead{Search \\ Cost} \\
\hline\hline
$\checkmark$ & $\times$ & 320 M & 74.2 & 112 \\
$\times$ & $\checkmark$ & 317 M & 74.2 & 188\\
$\times$ & $\times$ & 327 M & 73.8 & 307 \\
$\checkmark$ & $\checkmark$ & 325 M & 74.4 & 87  \\
\hline
\end{tabular}
\end{center}
\caption{Ablation study on each component of proposed method.}
\vspace{-0.5cm}
\label{tb:ab3}
\end{table}

Finally, we analyze the necessity of each component in our proposed data adapted pruning method. We partially disable each key component to examine the influence. As shown in Table \ref{tb:ab3}, our method full loaded largely reduces the search cost (from $307$ GPU hours to $87$ GPU hours) and yields the best searched architecture. This further proves the effectiveness of our method. 

\subsection{Transferring to Key-point Localization Task}

We further apply our method to the key-point localization task to demonstrate the generalization ability. Following the setup in simple baseline \cite{xiao2018simple}, we search a key-point localization architecture based on instance-level ground-truth. We modify our search space by attaching $3$ levels of de-convolution layers at the end, consisting de-convolution operators with $4 \times 4$ kernel and $2, 4, 8$ groups respectively. As shown in Table \ref{tb:kp}, our method is able to find a state-or-the-art architecture with significantly lower flops compared to manually designed methods~\cite{cpn_2018_CVPR,deeplabv3+,sun2019deep} with competing performance and significantly speedup previous best NAS method~\cite{nas_css}.

\begin{table}
\centering
\setlength{\tabcolsep}{1mm}{
\begin{tabular}{|l|c|c|c|c|c|c|c|c|c|c|}
\hline
\thead{Method} & \thead{Input size} & \thead{Search \\ Cost} & \thead{Params} & \thead{FLOPs} & \thead{AP} & \thead{AP$^{M}$} & \thead{AP$^{L}$} & \thead{AR}\\
\hline\hline
SimpleBaseline-ResNet50\cite{xiao2018simple} & 256 $\times$ 192 & Manual & 34.0 M & 8.90 G & 70.4 & 67.1 & 77.2 & 76.2\\
HRNet-W32\cite{sun2019deep} & 256 $\times$ 192 & Manual & 28.5 M & 7.10 G & 73.4 &  70.2 & 80.1 & 78.9\\
CPN-ResNet50\cite{cpn_2018_CVPR} & 256 $\times$ 192 & Manual & 27.0 M & 6.20 G & 69.4 & -- & -- & -- \\
DeepLab v3+\cite{deeplabv3+} & 256 $\times$ 192 & Manual & 5.8 M & -- & 66.8 & 64.1 & 70.7 & 70.0\\
\hline
NAS-CSS\cite{nas_css} & 256 $\times$ 192 & 192 & 2.9 M & -- & 65.9 & 63.1 & 70.0 & 69.3 \\
\hline
Ours & 256 $\times$ 192 & 30 & 10.9 M & 2.18 G & 68.4 & 65.5 & 74.4 & 75.7\\
\hline
\end{tabular}
}
\caption{Comparison to the state-of-the-art methods of key-point localization on COCO 2017 validation set.}
\vspace{-0.5cm}
\label{tb:kp}
\end{table}

\begin{figure}
\begin{center}
\includegraphics[width=0.9\linewidth]{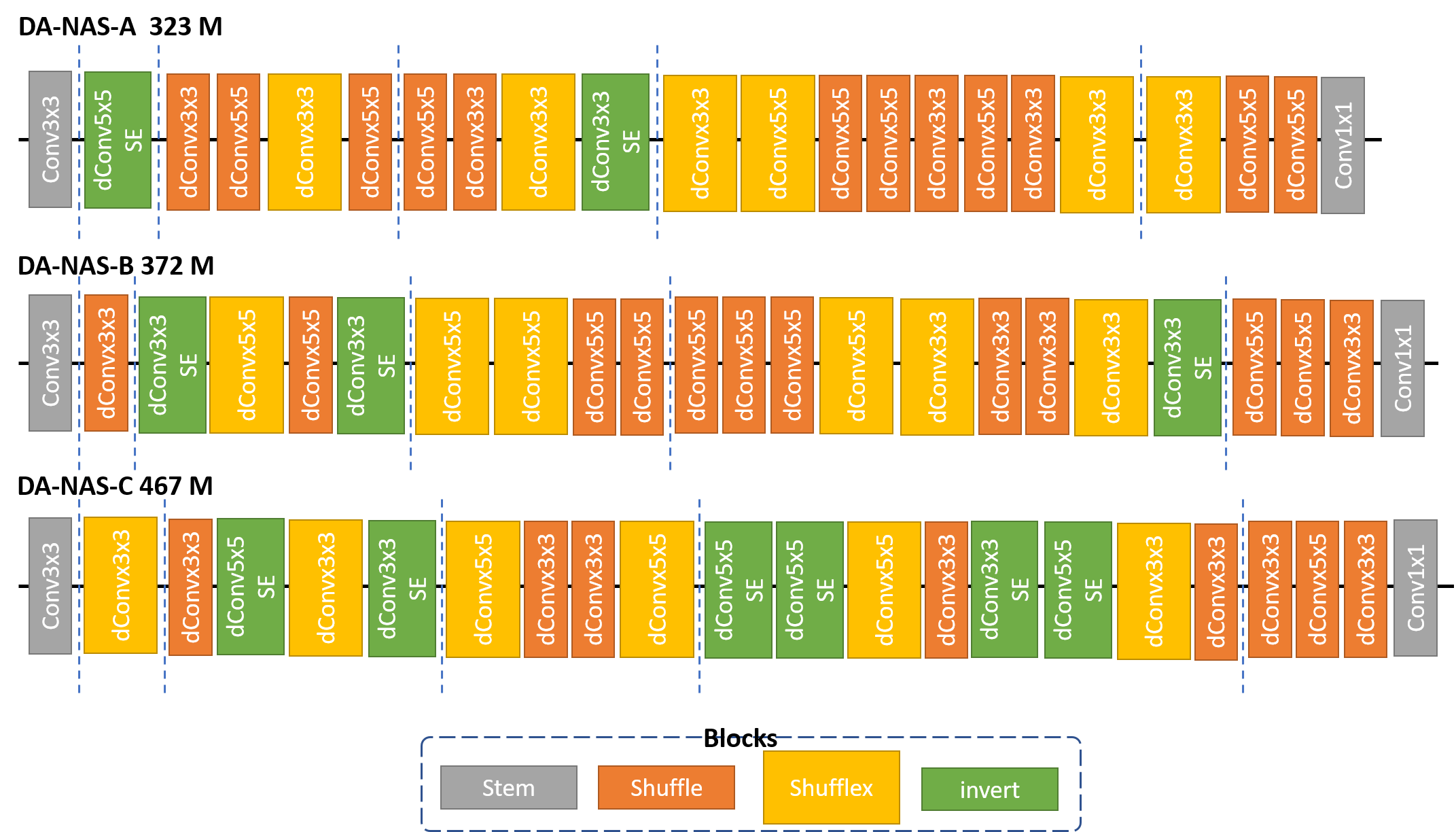}
\caption{Visualization of our best searched architectures. Network input flow is shown from left to right. Colored boxes indicate different blocks and texts in box indicate different operators. Blue dash lines after block indicate where the output resolutions reduce. The FLOPs of each architecture is marked at the beginning.}
\vspace{-0.5cm}
\label{fig:viz}
\end{center}
\end{figure}

\subsection{Visualization}
We visualize our searched architectures in Figure \ref{fig:viz} with two interesting findings. First, it is obvious to see that shuffle block is more cost efficient than inverted residual block. As the FLOPs constrain looses, the network architecture tends to have more inverted residual block to further boost the performance. Second, inverted residual block with squeeze-and-excitation component (SE) \cite{hu2018senet} is more likely to be placed at where number of channels increases or resolution of input reduces. This indicates that it is more effective to model inter-dependencies between different features channels. These findings are consistent with the statements from \cite{hu2018senet,ma2018shufflenet,howard2017mobilenets}. Thanks to the advantage of combining complex search spaces together, our method is capable of finding interesting properties of different blocks. 

\section{Conclusion}
In this paper, we present a novel data adapted pruning approach that largely speeds up neural architecture search. Inspired on the findings that network tends to learn easy categories first at early stage observed from network training, we propose to progressively utilize more data based on the easiness of classes, while pruning search space at the same time. Our strategy solves the conflicts between the requirement of large-scale data for fine-grained architecture search and the linearly increasing search cost, and makes the NAS practical in real-world task. Experiments show that our method can find state-of-the-art architecture with noticeable lower cost compared to popular methods. Our method is the first to combine data scheduling and search space pruning. Although a full understanding of the best setup of data and search space is not investigated in this paper, it opens a very interesting direction on how to effectively search based on data. Future work will focus on exploring the better data \& search space association and applying to much complicated task like object detection.

\bibliographystyle{splncs04}
\bibliography{egbib}

\begin{thebibliography}{10}
\providecommand{\url}[1]{\texttt{#1}}
\providecommand{\urlprefix}{URL }
\providecommand{\doi}[1]{https://doi.org/#1}

\bibitem{bilal2017convolutional}
Bilal, A., Jourabloo, A., Ye, M., Liu, X., Ren, L.: Do convolutional neural
  networks learn class hierarchy? IEEE Transactions on Visualization and
  Computer Graphics  \textbf{24}(1),  152--162 (2017)

\bibitem{cai2019once}
Cai, H., Gan, C., Han, S.: Once for all: Train one network and specialize it
  for efficient deployment. arXiv preprint arXiv:1908.09791  (2019)

\bibitem{cai2018proxylessnas}
Cai, H., Zhu, L., Han, S.: Proxylessnas: Direct neural architecture search on
  target task and hardware. arXiv preprint arXiv:1812.00332  (2018)

\bibitem{deeplabv3+}
Chen, L.C., Zhu, Y., Papandreou, G., Schroff, F., Adam, H.: Encoder-decoder
  with atrous separable convolution for semantic image segmentation. In:
  Ferrari, V., Hebert, M., Sminchisescu, C., Weiss, Y. (eds.) Computer Vision
  -- ECCV 2018. pp. 833--851. Springer International Publishing, Cham (2018)

\bibitem{cpn_2018_CVPR}
Chen, Y., Wang, Z., Peng, Y., Zhang, Z., Yu, G., Sun, J.: Cascaded pyramid
  network for multi-person pose estimation. In: The IEEE Conference on Computer
  Vision and Pattern Recognition (CVPR) (June 2018)

\bibitem{chen2019detnas}
Chen, Y., Yang, T., Zhang, X., Meng, G., Pan, C., Sun, J.: Detnas: Neural
  architecture search on object detection. arXiv preprint arXiv:1903.10979
  (2019)

\bibitem{chollet2017xception}
Chollet, F.: Xception: Deep learning with depthwise separable convolutions. In:
  Proceedings of the IEEE conference on computer vision and pattern
  recognition. pp. 1251--1258 (2017)

\bibitem{deng2009imagenet}
Deng, J., Dong, W., Socher, R., Li, L.J., Li, K., Fei-Fei, L.: Imagenet: A
  large-scale hierarchical image database. In: 2009 IEEE conference on computer
  vision and pattern recognition. pp. 248--255. Ieee (2009)

\bibitem{guo2019single}
Guo, Z., Zhang, X., Mu, H., Heng, W., Liu, Z., Wei, Y., Sun, J.: Single path
  one-shot neural architecture search with uniform sampling. arXiv preprint
  arXiv:1904.00420  (2019)

\bibitem{han2015deep}
Han, S., Mao, H., Dally, W.J.: Deep compression: Compressing deep neural
  networks with pruning, trained quantization and huffman coding. arXiv
  preprint arXiv:1510.00149  (2015)

\bibitem{he2016deep}
He, K., Zhang, X., Ren, S., Sun, J.: Deep residual learning for image
  recognition. In: Proceedings of the IEEE conference on computer vision and
  pattern recognition. pp. 770--778 (2016)

\bibitem{he2017channel}
He, Y., Zhang, X., Sun, J.: Channel pruning for accelerating very deep neural
  networks. In: Proceedings of the IEEE International Conference on Computer
  Vision. pp. 1389--1397 (2017)

\bibitem{howard2019mbnetv3}
Howard, A., Sandler, M., Chu, G., Chen, L., Chen, B., Tan, M., Wang, W., Zhu,
  Y., Pang, R., Vasudevan, V., Le, Q.V., Adam, H.: Searching for mobilenetv3.
  CoRR  \textbf{abs/1905.02244} (2019), \url{http://arxiv.org/abs/1905.02244}

\bibitem{howard2017mobilenets}
Howard, A.G., Zhu, M., Chen, B., Kalenichenko, D., Wang, W., Weyand, T.,
  Andreetto, M., Adam, H.: Mobilenets: Efficient convolutional neural networks
  for mobile vision applications. arXiv preprint arXiv:1704.04861  (2017)

\bibitem{hu2018senet}
Hu, J., Shen, L., Sun, G.: Squeeze-and-excitation networks (2018)

\bibitem{liang2019darts+}
Liang, H., Zhang, S., Sun, J., He, X., Huang, W., Zhuang, K., Li, Z.: Darts+:
  Improved differentiable architecture search with early stopping. arXiv
  preprint arXiv:1909.06035  (2019)

\bibitem{coco}
Lin, T.Y., Maire, M., Belongie, S., Hays, J., Perona, P., Ramanan, D.,
  Doll{\'a}r, P., Zitnick, C.L.: Microsoft coco: Common objects in context. In:
  Fleet, D., Pajdla, T., Schiele, B., Tuytelaars, T. (eds.) Computer Vision --
  ECCV 2014. pp. 740--755. Springer International Publishing, Cham (2014)

\bibitem{liu2019auto}
Liu, C., Chen, L.C., Schroff, F., Adam, H., Hua, W., Yuille, A.L., Fei-Fei, L.:
  Auto-deeplab: Hierarchical neural architecture search for semantic image
  segmentation. In: Proceedings of the IEEE Conference on Computer Vision and
  Pattern Recognition. pp. 82--92 (2019)

\bibitem{liu2018progressive}
Liu, C., Zoph, B., Neumann, M., Shlens, J., Hua, W., Li, L.J., Fei-Fei, L.,
  Yuille, A., Huang, J., Murphy, K.: Progressive neural architecture search.
  In: Proceedings of the European Conference on Computer Vision (ECCV). pp.
  19--34 (2018)

\bibitem{liu2018darts}
Liu, H., Simonyan, K., Yang, Y.: Darts: Differentiable architecture search.
  arXiv preprint arXiv:1806.09055  (2018)

\bibitem{ma2018shufflenet}
Ma, N., Zhang, X., Zheng, H.T., Sun, J.: Shufflenet v2: Practical guidelines
  for efficient cnn architecture design. In: Proceedings of the European
  Conference on Computer Vision (ECCV). pp. 116--131 (2018)

\bibitem{nekrasov2019fast}
Nekrasov, V., Chen, H., Shen, C., Reid, I.: Fast neural architecture search of
  compact semantic segmentation models via auxiliary cells. In: Proceedings of
  the IEEE Conference on Computer Vision and Pattern Recognition. pp.
  9126--9135 (2019)

\bibitem{nas_css}
Nekrasov, V., Chen, H., Shen, C., Reid, I.: Fast neural architecture search of
  compact semantic segmentation models via auxiliary cells. In: The IEEE
  Conference on Computer Vision and Pattern Recognition (CVPR) (June 2019)

\bibitem{pham2018efficient}
Pham, H., Guan, M.Y., Zoph, B., Le, Q.V., Dean, J.: Efficient neural
  architecture search via parameter sharing. arXiv preprint arXiv:1802.03268
  (2018)

\bibitem{real2019regularized}
Real, E., Aggarwal, A., Huang, Y., Le, Q.V.: Regularized evolution for image
  classifier architecture search. In: Proceedings of the AAAI Conference on
  Artificial Intelligence. vol.~33, pp. 4780--4789 (2019)

\bibitem{sandler2018mobilenetv2}
Sandler, M., Howard, A., Zhu, M., Zhmoginov, A., Chen, L.C.: Mobilenetv2:
  Inverted residuals and linear bottlenecks. In: Proceedings of the IEEE
  Conference on Computer Vision and Pattern Recognition. pp. 4510--4520 (2018)

\bibitem{simonyan2014very}
Simonyan, K., Zisserman, A.: Very deep convolutional networks for large-scale
  image recognition. arXiv preprint arXiv:1409.1556  (2014)

\bibitem{sun2019deep}
Sun, K., Xiao, B., Liu, D., Wang, J.: Deep high-resolution representation
  learning for human pose estimation. In: CVPR (2019)

\bibitem{Tan_2019_CVPR}
Tan, M., Chen, B., Pang, R., Vasudevan, V., Sandler, M., Howard, A., Le, Q.V.:
  Mnasnet: Platform-aware neural architecture search for mobile. In: The IEEE
  Conference on Computer Vision and Pattern Recognition (CVPR) (June 2019)

\bibitem{wu2019fbnet}
Wu, B., Dai, X., Zhang, P., Wang, Y., Sun, F., Wu, Y., Tian, Y., Vajda, P.,
  Jia, Y., Keutzer, K.: Fbnet: Hardware-aware efficient convnet design via
  differentiable neural architecture search. In: Proceedings of the IEEE
  Conference on Computer Vision and Pattern Recognition. pp. 10734--10742
  (2019)

\bibitem{xiao2018simple}
Xiao, B., Wu, H., Wei, Y.: Simple baselines for human pose estimation and
  tracking. In: European Conference on Computer Vision (ECCV) (2018)

\bibitem{xie2018snas}
Xie, S., Zheng, H., Liu, C., Lin, L.: {SNAS}: stochastic neural architecture
  search. In: International Conference on Learning Representations (2019),
  \url{https://openreview.net/forum?id=rylqooRqK7}

\bibitem{yan2019hm}
Yan, S., Fang, B., Zhang, F., Zheng, Y., Zeng, X., Zhang, M., Xu, H.: Hm-nas:
  Efficient neural architecture search via hierarchical masking. In:
  Proceedings of the IEEE International Conference on Computer Vision
  Workshops. pp.~0--0 (2019)

\bibitem{zhang2018shufflenet}
Zhang, X., Zhou, X., Lin, M., Sun, J.: Shufflenet: An extremely efficient
  convolutional neural network for mobile devices. In: Proceedings of the IEEE
  Conference on Computer Vision and Pattern Recognition. pp. 6848--6856 (2018)

\bibitem{zhou2017incremental}
Zhou, A., Yao, A., Guo, Y., Xu, L., Chen, Y.: Incremental network quantization:
  Towards lossless cnns with low-precision weights. arXiv preprint
  arXiv:1702.03044  (2017)

\bibitem{nasnet_2018_CVPR}
Zoph, B., Vasudevan, V., Shlens, J., Le, Q.V.: Learning transferable
  architectures for scalable image recognition. In: The IEEE Conference on
  Computer Vision and Pattern Recognition (CVPR) (June 2018)

\end{thebibliography}

\end{document}